\documentclass[runningheads]{llncs}
\usepackage{amsmath,amssymb} 
\usepackage{color}
\usepackage[width=122mm,left=12mm,paperwidth=146mm,height=193mm,top=12mm,paperheight=217mm]{geometry}

\usepackage[utf8]{inputenc} 
\usepackage[T1]{fontenc}    
\usepackage{booktabs}       
\usepackage{amsfonts}       
\usepackage{nicefrac}       
\usepackage{microtype}      
\usepackage{color}
\usepackage{colortbl}
\usepackage{tabularx}
\usepackage{subcaption}
\usepackage{arydshln}
\usepackage{array,multirow}
\usepackage{xfrac}
\usepackage{bm}
\usepackage{soul}
\usepackage[export]{adjustbox}
\usepackage{graphicx}
\usepackage{verbatim}
\usepackage{mwe}
\usepackage[shortcuts]{extdash}
\usepackage[toc,page]{appendix}
\usepackage[ruled,lined,linesnumberedhidden]{algorithm2e}
\usepackage{enumitem}
\usepackage{bbold}
\usepackage[normalem]{ulem}
\usepackage[dvipsnames]{xcolor}
\usepackage{bbm}
\usepackage{ulem}
\usepackage{float}

\newcolumntype{C}[1]{>{\centering\let\newline\\\arraybackslash\hspace{0pt}}m{#1}}
\newcolumntype{K}[1]{>{\centering\arraybackslash}p{#1}}

\newcommand{\superscript}[1]{$^{\text{#1}}$}

\newcommand{\rb}{\rotatebox{90}}%

\newcommand{\eg}{\textit{e}.\textit{g}.}
\newcommand{\etal}{\textit{et al}.}

\newcommand{\beginsupplement}{%
        \setcounter{table}{0}
        \renewcommand{\thetable}{\Alph{table}}%
        \setcounter{figure}{0}
        \renewcommand{\thefigure}{\Alph{figure}}%
        \setcounter{section}{0}
        \renewcommand{\thesection}{\Alph{section}}%
     }

\begin{document}

\pagestyle{headings}
\mainmatter
\def\ECCV18SubNumber{2325}  

\title{Attentive Semantic Alignment \\ with Offset-Aware Correlation Kernels} 

\titlerunning{Attentive Semantic Alignment with Offset-Aware Correlation Kernels}

\authorrunning{P. H. Seo, J. Lee, D. Jung, B. Han and M. Cho}

\author{Paul Hongsuck Seo\inst{1} \and Jongmin Lee\inst{1} \and\\ Deunsol Jung\inst{1} \and Bohyung Han\inst{2} \and Minsu Cho\inst{1}}
\institute{Pohang University of Science and Technology (POSTECH), Korea \and
           Dept. of ECE \& ASRI, Seoul National University, Korea \\
           \email{\{hsseo, ljm1121, hesedjds, mscho\}@postech.ac.kr} 
           \email{ bhhan@snu.ac.kr}}

\maketitle


\begin{abstract}

Semantic correspondence is the problem of establishing correspondences across images depicting different instances of the same object or scene class. 
One of recent approaches to this problem is to estimate parameters of a global transformation model that densely aligns one image to the other. Since an entire correlation map between all feature pairs across images is typically used to predict such a global transformation, noisy features from different backgrounds, clutter, and occlusion distract the predictor from correct estimation of the alignment. This is a challenging issue, in particular, in the problem of semantic correspondence where a large degree of image variations is often involved. 
In this paper, we introduce an attentive semantic alignment method that focuses on reliable correlations, filtering out distractors.
For effective attention, we also propose an offset-aware correlation kernel that learns to capture translation-invariant local transformations in computing correlation values over spatial locations.
Experiments demonstrate the effectiveness of the attentive model and offset-aware kernel, and the proposed model combining both techniques achieves the state-of-the-art performance.

\end{abstract}


\section{Introduction}
\label{sec:introduction}

Semantic correspodence is the problem of establishing correspondences across images depicting different instances
of the same object or scene class. 
Compared to conventional correspondence tasks handling pictures of the same scene, such as stereo matching~\cite{hosni2013fast,okutomi1993multiple} and motion estimation~\cite{dosovitskiy2015flownet,weinzaepfel2013deepflow,revaud2016deepmatching}, 
  the problem of semantic correspondence involves substantially larger changes in appearance and spatial layout, thus remaining very challenging.
For this reason, traditional approaches based on hand-crafted features such as SIFT~\cite{lowe2004distinctive,liu2016sift} and HOG~\cite{dalal2005histograms,ham2016proposal,taniai2016joint} 
do not produce satisfactory results on this problem due to lack of high-level semantics in local feature representations.

\begin{figure}[t]
    \centering
    \includegraphics[width=1.0\linewidth]{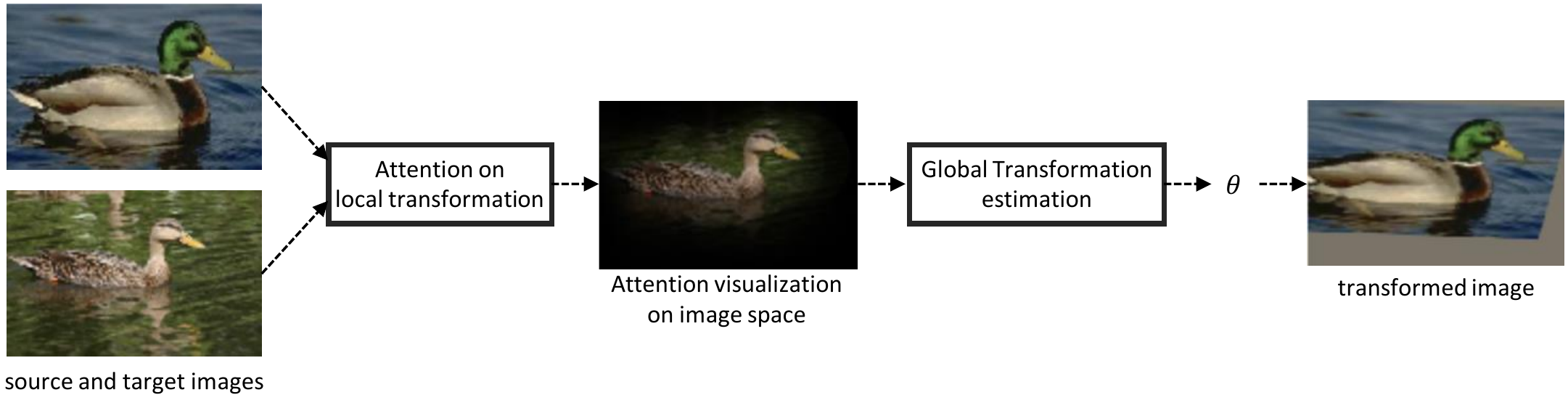}
    \caption{
    The proposed attentive semantic alignment. 
    Our model estimates dense correspondences of objects by predicting a set of global transformation parameters via an attention process. The attention process spatially focuses on the reliable local transformation features, filtering out irrelevant backgrounds and clutter. 
    }
    \label{fig:geo_matching}
\end{figure}

While previous approaches to the problem focus on introducing an effective spatial regularizer in matching~\cite{liu2016sift,kim2013deformable,ham2016proposal},  
recent convolutional neural networks have advanced this area 
by learning high-level semantic features~\cite{choy2016universal,rocco2017convolutional,han2017scnet,ufer2017deep,kim2017dctm,kim2017fcss,novotny2017anchornet}.
One of the main approaches~\cite{rocco2017convolutional} is to estimate parameters of a global transformation model that densely aligns one image to the other. 
In contrast to other approaches, it casts the whole correspondence problem for all individual features into a simple regression problem with a global transformation model, thus predicting dense correspondences through the efficient pipeline. 
On the other hand, however, the global alignment approach may be easily distracted; 
An entire correlation map between all feature pairs across images is used to predict such a global transformation, and thus noisy features from different backgrounds, clutter, and occlusion, may distract the predictor from correct estimation of the alignment. This is a challenging issue, in particular, in the problem of semantic correspondence where a large degree of image variations is often involved. 

In this paper, we introduce an attentive semantic alignment method that focuses on reliable correlations, filtering out distractors as shown in Fig.~\ref{fig:geo_matching}.
For effective attention, we also propose an offset-aware correlation kernel that learns to capture translation-invariant local transformations in computing correlation values over spatial locations.
The resultant feature map of offset-aware correlation (OAC) kernels is computed from two input features, where each activation of the feature map represents how smoothly a source feature is transformed spatially to the target feature map. This use of OAC kernels greatly improves a subsequent attention process. 
Experiments demonstrate the effectiveness of the attentive model and offset-aware kernel, and the proposed model combining both techniques achieves the state-of-the-art performance.

Our contribution in this work is threefold:
\begin{itemize}
    \item[$\bullet$] The proposed algorithm incorporates an attention process to estimate a global transformation from a set of inconsistent and noisy local transformations for semantic image alignment.
    \item[$\bullet$] We introduce offset-aware correlation kernels to guide the network in capturing local transformations at each spatial location effectively, and employ the kernels to compute feature correlations between two images for better representation of semantic alignment.
    \item[$\bullet$] The proposed network with the attention module and offset-aware correlation kernels achieves the state-of-the-art performances on semantic correspondence benchmarks.
\end{itemize}

The rest of the paper is organized as follows. 
We overview the related work in Section~\ref{sec:rel_work}.
Section~\ref{sec:net_arch} describes our proposed network with the attention process and the offset-based correlation kernels.
Finally, we show the experimental results of our method and conclude the paper in Section~\ref{sec:exp} and \ref{sec:conclusion}.

\section{Related Work}
\label{sec:rel_work}

Most approaches to semantic correspondence are based on dense matching of local image features. 
Early methods extract local features of patches using hand-crafted feature descriptors~\cite{yang2014daisy} such as SIFT~\cite{kim2013deformable,liu2016sift,hur2015generalized,bristow2015dense} and HOG~\cite{ham2016proposal,berg2005shape,taniai2016joint,yang2017object}.
In spite of some success, the lack of high-level semantics in the feature representation makes the approaches suffer from non-rigid deformation and large appearance changes of objects.
While such challenges have been mainly investigated in the area of graph-based image matching~\cite{berg2005shape,cho2010reweighted,duchenne2011graph,cho2013learning}, recent methods~\cite{ufer2017deep,kim2017dctm,kim2017fcss,novotny2017anchornet,zagoruyko2015learning,zbontar2015computing,han2015matchnet,long2014convnets,zhou2016learning,kanazawa2016warpnet} rely on deep neural networks to extract high-level features of patches for robust matching.
More recently, Han \etal~\cite{han2017scnet} propose a deep neural network that learns both a feature extractor and a matching model for semantic correspondence. 
In spite of these developments, all these approaches detect correspondences by matching patches or region proposals based on their local features. In contrast, 
Rocco \etal~\cite{rocco2017convolutional} propose a global transformation estimation method that is the most relevant work to ours. Their model in \cite{rocco2017convolutional} predicts the transformation parameters from a correlation map obtained by computing correlations of every pair of features in source and target feature maps.
Although this model is similar to ours in that it estimates the global transformation based on correlations of feature pairs, our model is distinguished by the attention process suppressing irrelevant features and the use of the OAC kernels constructing local transformation features.
%

There are some related studies on other tasks using feature correlations such as optical flow estimation~\cite{dosovitskiy2015flownet} and stereo matching~\cite{zbontar2016stereo,luo2016efficient}.
Dosovitskiy \etal\cite{dosovitskiy2015flownet} use correlations between features of two video frames to estimate optical flow, while Zbontar \etal\cite{zbontar2016stereo} and Luo \etal\cite{luo2016efficient} extract feature correlations from patches of images for stereo matching.
Although all these methods utilize the correlations, they extract correlations from features in a limited set of candidate regions.
Moreover, unlike ours, they do not explore the attentive process and the offset-based correlation kernels.

Lately, attention models have been widely explored for various tasks with multi-modal inputs such as image captioning~\cite{xu2015show,mun2017text}, visual question answering~\cite{xu2016ask,yang2016stacked}, attribute prediction~\cite{seo2016progressive} and machine translation~\cite{bahdanau2015neural,luong2015effective}.
In these studies, models attend to the relevant regions referred and guided by another modality such as language, while the proposed model attends based on a self-guidance.
Noh \etal~\cite{noh2017large} use an attention process for image retrieval to extract deep local features, where the attention is obtained from the features themselves as in our work.



\section{Deep Attentive Semantic Alignment Network}
\label{sec:net_arch}

\begin{figure}[t]
\centering
\includegraphics[width=1\linewidth]{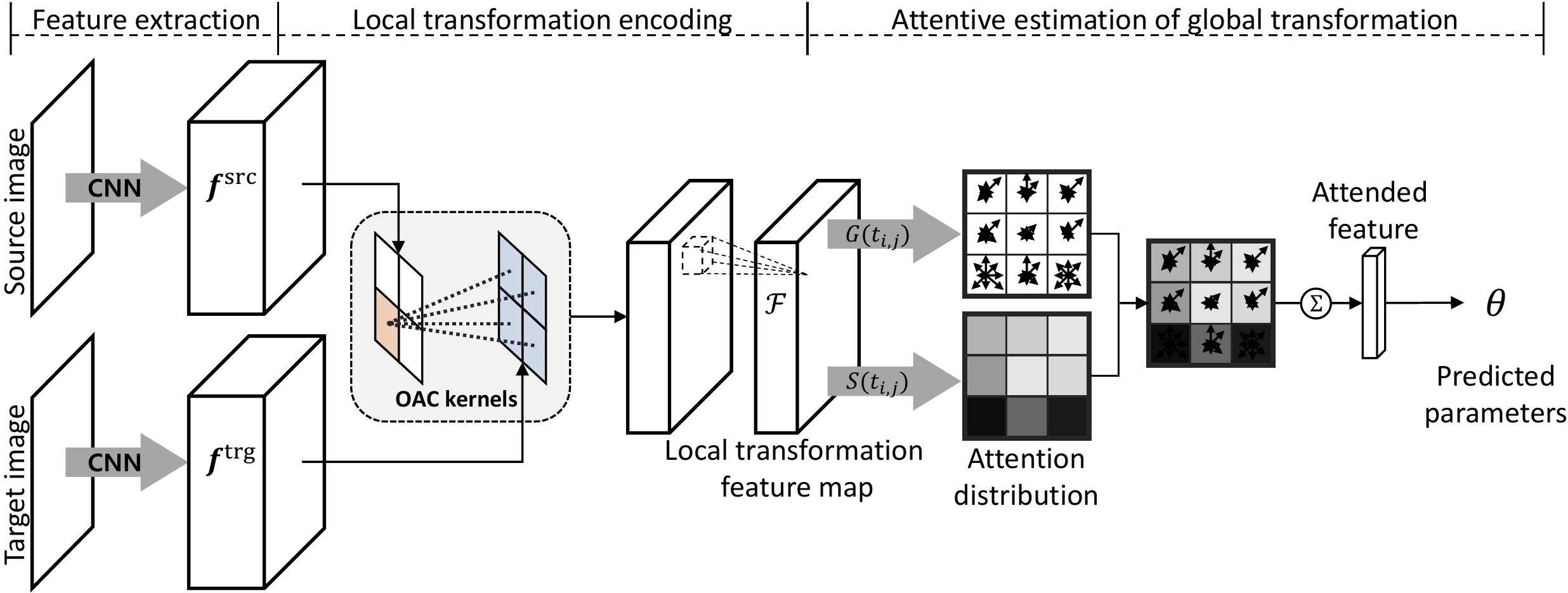}
\caption{
Overall architecture of the proposed network.
It consists of three main components: feature extractor, local transformation encoder, and attentive global transformation estimator. For details, see text. 
}
\label{fig:net_arch}
\end{figure}

We propose a deep neural network architecture for semantic alignment incorporating an attention process with a novel offset-aware correlation kernel.
Our network takes as inputs two images and estimates a set of global transformation parameters using three main components: feature extractor, local transformation encoder, and attentive global transformation estimator as presented in Fig.~\ref{fig:net_arch}.
We describe each of these components in details.

\subsection{Feature extractor}
Given source and target images, we first extract their image feature maps $\bm{f}^\mathrm{src},\bm{f}^\mathrm{trg} \in \mathbbm{R}^{D\times H\times W}$ using a fully convolutional image feature extractor, where $H$ and $W$ are height and width of input images, respectively.
We use a VGG-16~\cite{simonyan2015very} model pretrained on ImageNet~\cite{deng2009imagenet} and extract features from its \texttt{pool4} layer.
We share the weights of the feature extractor for both source and target images.
Input images are resized into $240\times240$ and fed to the feature extractor resulting in $15\times15$ feature maps with $512$ channels.
After extracting the features, we normalize them using $L_2$ norm.

\subsection{Local transformation encoder}
Given source and target feature maps from the feature extractor, the model encodes local transformations of the source features with respect to the target feature map.
The encoding is given by introducing a novel offset-aware correlation (OAC) kernel, which facilitates to overcome limitations of conventional correlation layers~\cite{rocco2017convolutional}.
We briefly describe details of the correlation layer including its limitations and discuss the proposed OAC kernel.

\subsubsection{Correlation layer}

The correlation layer computes correlations of all pairs of features from the source and the target images~\cite{rocco2017convolutional}.
Specifically, the correlation layer takes two feature maps as its inputs and constructs a correlation map $\bm{c} \in \mathbbm{R}^{HW \times H \times W}$, which is given by
\begin{equation}
    c_{i,j} = {f^\mathrm{src}_{i,j}}^\top \hat{\bm{f}}^\mathrm{trg},
\end{equation}
where $c_{i,j} \in \bm{c}$ is a $HW$ dimensional correlation vector at a spatial location $(i, j)$, $f_{i,j}^\mathrm{src} \in \bm{f}^\mathrm{src}$ is a feature vector at a location $(i, j)$ of the source image, and $\hat{\bm{f}}^\mathrm{trg} \in \mathbbm{R}^{D\times HW}$ is a spatially flattened feature map of $\bm{f}^\mathrm{trg}$ of the target image.
In other words, each correlation vector $c_{i,j}$ consists of correlations between a single source feature $f_{i,j}^\mathrm{src}$ and all target features of $\bm{f}^\mathrm{trg}$.
Although each element of a correlation vector maintains the correspondence likelihood of a source feature onto a certain location in the target feature map, the order of elements in the correlation vector is based on the absolute coordinates of individual target features regardless of the source feature location. This means that decoding the local displacement of the source feature requires not only the vector itself but also the spatial location of the source feature.
For example, consider a correlation vector $c_{i,j}=[1, 0, 0, 0]^\top$ between $2\times 2$ feature maps, each element of which is the correlation of $f^\mathrm{src}_{i,j}$ with $f^\mathrm{trg}_{0,0}$, $f^\mathrm{trg}_{0,1}$, $f^\mathrm{trg}_{1,0}$ and $f^\mathrm{trg}_{1,1}$. The displacement represented by the vector varies with the coordinate of the source feature $(i,j)$. When $(i,j)=(0,0)$, it indicates that the source feature $f^\mathrm{src}_{0,0}$ remains at the same location $(0,0)$ in the target feature map. When $(i,j)=(0,1)$, it implies that $f^\mathrm{src}_{0,1}$ is moved to the left of its original location in the target feature map.

Given a correlation map, decoding the local displacement of a source feature requires incorporating the offset information from the source feature to individual target features. And, this local process is crucial for subsequent spatial attention process in the next section. Therefore, we first introduce an offset-aware correlation kernel that utilizes the offset of features during the kernel application.


\begin{figure}[t]
\centering
\begin{subfigure}[b]{0.49\linewidth}
\centering
\includegraphics[width=0.8\linewidth]{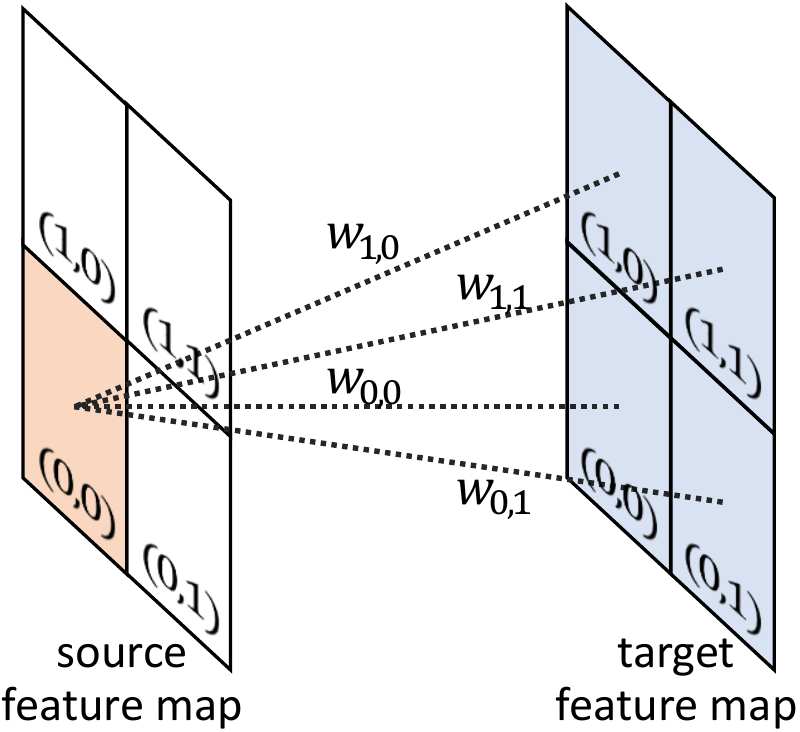}
\subcaption{kernel alignment for $h_{0,0}$}
\label{fig:kernels_00}
\end{subfigure}
\begin{subfigure}[b]{0.49\linewidth}
\centering
\includegraphics[width=0.8\linewidth]{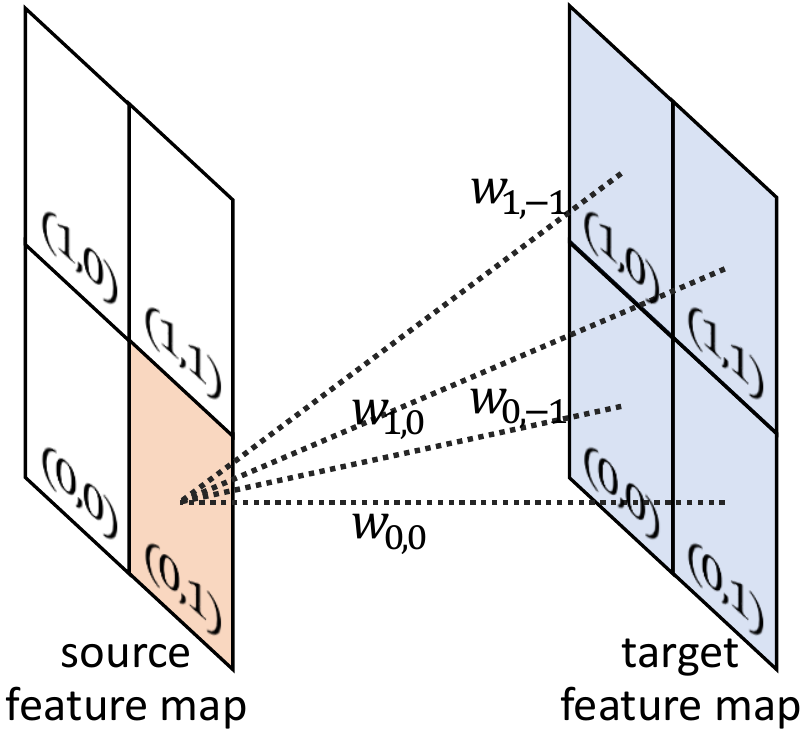}
\subcaption{kernel alignment for $h_{0,1}$}
\label{fig:kernels_01}
\end{subfigure}
\caption{
Offset-aware correlation kernel at different source locations: (a) at $(0,0)$ and (b) at $(0,1)$.
Each dotted line connects source and target features to compute correlation, and $w_{i,j}$ represents a kernel weight for the dotted line.
Note that kernel weights are associated with different correlation pairs when source locations vary.
}
\label{fig:kernels}
\end{figure}

\subsubsection{Offset-aware correlation kernels}
Similarly to the correlation layer, our OAC kernels also take two input feature maps and utilize correlations of all feature pairs between these feature maps.
The kernels naturally capture the displacement of a source feature in the target feature map by aligning kernel weights based on the offset between the source and target features for each correlation as illustrated in Fig.~\ref{fig:kernels}.
Formally speaking, an OAC kernel captures feature displacement of a source feature $f^\mathrm{src}_{i,j}$ by
\begin{align}
    h_{i, j}^{(n)} &=\sum_{k=1}^{H}{\sum_{l=1}^{W}{w_{i-k, j-l}^{(n)} c_{i,j;k,l}}} \\
    &=\sum_{k=1}^{H}{\sum_{l=1}^{W}{w_{i-k, j-l}^{(n)} {f^\mathrm{src}_{i,j}}^\top f^\mathrm{trg}_{k,l}}},
\end{align}
where $h^{(n)}_{i,j}$ is the kernel output with the kernel index $n$, $c_{i,j;k,l}$ is the correlation between a source feature $f^\mathrm{src}_{i,j}$ and a target feature $f^\mathrm{trg}_{k,l}$, and $\Phi^{(n)}= \{w^{(n)}_{s,t}\}$ is a set of the kernel weights.
Note that the kernel weights are indexed by offset between the source and target features, and shared for correlations of any feature pair with the same offset.
For example, in Fig.~\ref{fig:kernels_00}, $w_{0,0}$ is associated with the target feature at $(0,0)$ because the source location is $(0,0)$.
The same weight $w_{0,0}$ is associated with the target feature at $(0,1)$ when the source location is $(0,1)$ as in Fig.~\ref{fig:kernels_01} because the offset between these features is $(0,0)$.
Also note that each kernel output $h^{(n)}_{i,j}$ at a location $(i,j)$ captures the displacement of its corresponding source feature $f^\mathrm{src}_{i,j}$ at the same location.

While a proposed kernel captures a single aspect of feature displacement, a set of the proposed kernels produce a dense feature representation of feature displacement for each source feature.
We use $128$ OAC kernels resulting in a feature displacement map $\bm{h} \in \mathbbm{R}^{128\times15\times15}$ encoding the displacement of each source feature.
We set ReLU as the activation functions of the kernel outputs, and compute normalized correlations in OAC kernels since normalization further improves the scores as observed in \cite{rocco2017convolutional}.

In practice, the proposed OAC kernels are implemented by two sub-procedures.
We first compute the normalized correlation map reordered based on the offsets between the locations of the source and target features.
In this reordered correlation map, every correlation with the same relative displacement is arranged in the same channel.
This reordering results in $(2H-1)(2W-1)$ possible offsets and thus the size of the output tensor becomes $(2H-1)(2W-1) \times H \times W$ where many of the values are zeros due to non-existing pairs for some offsets. 
Then, we use a $1\times1$ convolutional layer to compute the dense feature representation from the raw displacement information captured in the reordered correlation map.
Note that this process significantly reduces the number of channels by compactly encoding various aspects of the local displacements into dense representations.



\subsubsection{Encoding local transformation features}
Since the feature displacement map conveys the movement of each source feature independently, each feature alone is not sufficient to predict the global transformation parameters.
To allow the network predicts the global transformation from local features in the attention process, we construct a local transformation feature map by combining spatially adjacent feature displacement information captured by $\bm{h}$.
That is, the proposed network feeds the feature displacement map $\bm{h}$ to a $7\times7$ convolution layer with $128$ output channels applied without padding.
This convolution layer results in a local transformation feature map $\mathcal{F}\in\mathbbm{R}^{128\times9\times9}$.
Note that each feature $t_{i,j} \in \mathcal{F}$ captures transformations occurred in a local region. 
We utilize this local transformation feature map to predict the global transformation through an attention process.

\begin{figure}[t]
    \centering
    \includegraphics[width=1\linewidth]{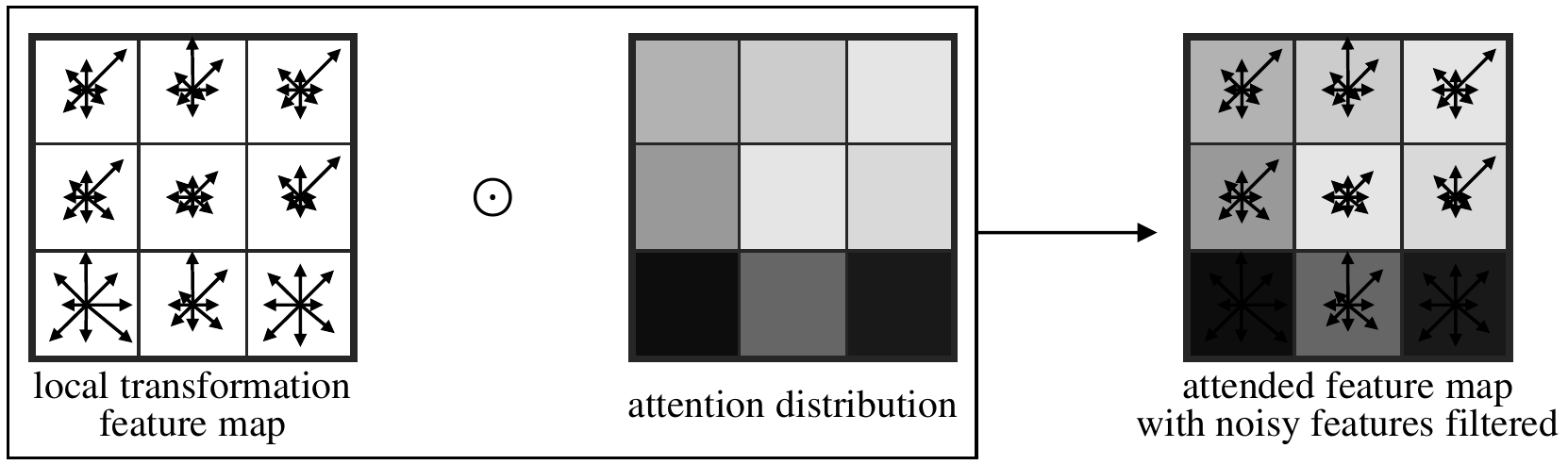}
    \caption{
    Illustration of attention process. 
    Noisy features in local transformation feature map are filtered by assigning lower probabilities to these locations.
    Arrows in boxes of local transformation feature map demonstrate features encoding local transformations, and grayscale colors in attention distribution represent magnitudes of probabilities where brighter colors mean higher probabilities.}
    \label{fig:att}
\end{figure}

\subsection{Attentive global transformation estimator}
\label{sub:att}
After local transformation encoding, a set of global transformation parameters is estimated with an attention process.
Given a local transformation feature map $\mathcal{F}\in \mathbbm{R}^{\hat{D}\times\hat{H}\times\hat{W}}$ extracted by OAC kernels with a convolution layer, the network focuses on reliable local transformation features by filtering out distracting regions as depicted in Fig.~\ref{fig:att} to predict the parameters from the aggregation of those features. 
Although a feature map $\mathcal{F}$ gives sufficient information to predict the global transformation from source to target, local transformation features extracted from a real image pair is noisy due to image variations such as background clutter and intra-class variations.
Therefore, we propose a model that suppresses unreliable features by the attention process and extracts an attended feature vector that summarizes local transformations from all reliable locations to estimate an accurate global transformation.
In other words, the model computes an attended transformation feature $\tau^\mathrm{att}$ by
\begin{align}
    \tau^\mathrm{att} 
    = \sum_{i=1}^{\hat{H}}{\sum_{j=1}^{\hat{W}}{\alpha_{i,j}G(t_{i, j})}},
\label{eq:attention}
\end{align}
where $G:\mathbbm{R}^{\hat{D}}\rightarrow \mathbbm{R}^{D'}$ is a projection function of $t_{i,j}$ into a $D'$ dimensional vector space and $\bm{\alpha} = \{\alpha_{i,j}\}$ is an attention probability distribution over feature map.
The model computes the attention probabilities by
\begin{equation}
    \alpha_{i,j} = \frac{\exp\left({S(t_{i,j})}\right)}{\sum_{k=1}^{\hat{H}}{\sum_{l=1}^{\hat{W}}{\exp\left({S(t_{k,l})}\right)}}},\label{eq:att_prob}
\end{equation}
where $S:\mathbbm{R}^{\hat{D}}\rightarrow\mathbbm{R}$ is an attention score function producing a single scalar given a local transformation feature.
Note that the model learns to suppress noisy features by assigning low attention scores and reducing their contribution to the attended feature.

Once the attended feature $\tau^\mathrm{att}$ over all local transformations is obtained, we compute the global transformation $\theta \in \mathbbm{R}^Q$ by a simple matrix-vector multiplication as
%
\begin{equation}
    \theta = W\tau^\mathrm{att}, \label{eq:prediction}
\end{equation}
where $W \in \mathbbm{R}^{Q\times{D'}}$ is a weight matrix for linear projection of the attended feature $\tau^\mathrm{att}$.

In summary, we first compute local transformation between two images and perform a nonlinear embedding using a projection function $G(\cdot)$.
The embedded vector is weighted by spatial attention to compute an attended feature $\tau^\mathrm{att}$ as shown in Eq.~\eqref{eq:attention}.
The global transformation vector is obtained by linear projection of the attended feature, which is parametrized by a matrix as presented in Eq.~\eqref{eq:prediction}.


We use multi-layer perceptrons (MLPs) for $G$ and $S$ in Eq.~\eqref{eq:attention} and \eqref{eq:att_prob}.
$G$ is a two-layer MLP with $128$ hidden and output ReLU activations.
Since the feature representations produced by $G$ is directly used for the final estimation as a linear mapping in Eq.~\eqref{eq:prediction}, we additionally concatenate $5$-dimensional index embedding to the feature $t_{i,j}\in\mathcal{F}$ to better estimate the global transformation from local transformation features.
While $S$ is another two-layer MLP with $64$ hidden ReLU activations, its output is a scalar without non-linearity; 
this is due to the application of softmax normalization outside $S$.
Note that we do not use the index embedding to avoid strong biases of attentions on certain regions.
Since $G$ and $S$ are applied to all feature vectors across the spatial dimensions, we implement them by multiple $1\times1$ convolutions with batch normalizations.

\subsubsection{Network training}
We build two of the proposed networks with different parametric global transformations: affine and thin-plate spline (TPS) transformations.
To train the network, we adapt the average transformed grid distance loss proposed in~\cite{rocco2017convolutional}, which indirectly measures the distance from the predicted transformation parameters $\theta$ to the ground-truth transformation parameters $\theta_\mathrm{GT}$.
Given $\theta$ and $\theta_\mathrm{GT}$, the transformed grid distance $\mathrm{TGD}(\theta, \theta_\mathrm{GT})$ is obtained by
\begin{equation}
    \mathrm{TGD}(\theta, \theta_\mathrm{GT}) = \frac{1}{|\mathcal{G}|}\sum_{g \in \mathcal{G}}{d\left(\mathcal{T}_\theta\left(g\right), \mathcal{T}_{\theta_\mathrm{GT}}\left(g\right)\right)^2}
\end{equation}
where $\mathcal{G}$ is a set of points in a regular grid, $\mathcal{T}_\theta$ is the transformation parameterized by $\theta$ and $d(\cdot)$ is a distance measure.
We minimize the average TGD of training examples to train the network.
Since every operation within the proposed network is differentiable, the network is trainable end-to-end using a gradient-based optimization algorithm.
We use ADAM~\cite{kingma2014adam} with initial learning rate of $2 \times 10^{-4}$ and batch size of $32$ for $50$ epochs.
During training, the pretrained feature extractor is fixed and only the other parts of the network are finetuned.


\section{Experiments}
\label{sec:exp}

We evaluate the proposed method on public benchmarks for semantic correspondence estimation.
The experiments demonstrate that the proposed attentive method and OAC kernels are effective in semantic alignment, substantially improving the baseline models.
The codes are publicly released at \url{http://cvlab.postech.ac.kr/research/A2Net/}.

\subsection{Experimental settings}

\subsubsection{Training with self-supervision}
While the loss function requires the full supervision of $\theta_\mathrm{GT}$, it is very expensive or even impractical to collect exact ground-truth transformation parameters for non-rigid objects involving intra-class variations.
Therefore, it is hard to scale up to numerous instances and classes, restricting generalization.
For example, the largest annotation dataset at this time, PF-PASCAL, only contains total 1,351 image pairs from 20 classes, and furthermore their dense annotations are extrapolated from sparse keypoints, thus being not fully exact.
To work around this problem, we adopt the self-supervised learning for semantic alignment, which is free from the burden of any manual annotation, is an appealing alternative introduced in \cite{rocco2017convolutional}.
In this framework, given a public image dataset $\mathcal{D}$ without any annotations, we synthetically generate a training example $(I_\mathrm{src}, I_\mathrm{trg})$ by randomly sampling an image $I_{src}$ from $\mathcal{D}$ and computing a transformed image $I_{trg}$ by applying a random transformation $\theta_\mathrm{GT}$.
We also use mirror padding and center cropping following \cite{rocco2017convolutional} to avoid border artifacts.
The synthetic image pairs generated by this process are annotated with the ground-truth transformation parameters $\theta_\mathrm{GT}$ allowing us to train the network with full supervision.
Note that, however, this training scheme can be considered unsupervised since no annotated real dataset is used during training.

For the synthetic dataset generation, we use PASCAL VOC 2011~\cite{pascal-voc-2011}, and build two variations of training datasets with either affine or TPS transformation each for its corresponding network.
A set of PASCAL VOC images is kept separate to generate another set of synthetic examples for validation and the best performing models on the validation set is evaluated.

\subsubsection{Evaluation}
Two public benchmarks called PF-WILLOW and PF-PASCAL~\cite{ham2016proposal} are used for the evaluation.
PF-WILLOW consists of about $900$ image pairs generated from $100$ images of 5 object classes. PF-PASCAL contains 1351 image pairs of 20 object classes.
Each image pair in both datasets contains different instances of the same object class such as ducks and motorbikes, \eg, left two images in Fig.~\ref{fig:geo_matching}.
The objects in these datasets have large intra-class variations and many background clutters making the task more challenging.
The image pairs of both PF-WILLOW and PF-PASCAL are annotated with sparse key points that establishe correspondences between two images.
Following the standard evaluation metric, the probability of correct keypoint (PCK)~\cite{yang2013articulated} of these benchmarks, our goal is to correctly transform the key points in the source image to their corresponding ones in the target image.
A transformed source key point is considered correct if its distance to its corresponding target key point is less than $\alpha\cdot\mathrm{max}(h, w)$, where $\alpha=0.1$, and $h$ and $w$ are height and width of the object bounding box.
Formally, PCK of a proposed model $\mathcal{M}$ is measured by
\begin{equation}
    \mathrm{PCK}(\mathcal{M})= \frac{\sum\limits_{i=1}^{N}\sum\limits_{(p_\mathrm{s}, p_\mathrm{t})\in \mathcal{P}_i}\mathbb{1}\left[d\left(\mathcal{T}_{\theta_i}\left(p_\mathrm{s}\right), p_\mathrm{t}\right) < \alpha\cdot \mathrm{max}(h, w)\right]}{\sum\limits_{i=1}^{N}{|\mathcal{P}_i|}},
\end{equation}
where $N$ is the total number of image pairs, $\mathcal{P}_i$ is a set of source and target key point pairs $(p_\mathrm{s},p_\mathrm{t})$ for $i$\superscript{th} example, $\theta_i$ is predicted transformation, and $\mathbb{1}$ is the indicator function which returns $1$ if the expression inside brackets is true and $0$ otherwise.

We evaluate three different versions of the proposed model as in~\cite{rocco2017convolutional}.
The first two versions are the models with different transformations: affine and TPS transformations.
The other version sequentially merges these two models.
That is, the input image pair are first fed to the network with affine transformation, and the image pair transformed by its out is then fed to the network with TPS transformation.


\begin{table}[t]
\centering
\caption{
Experimental results on PF-WILLOW and PF-PASCAL. 
PCK is measured with $\alpha=0.1$.
Scores for other models are brought from~\cite{ham2016proposal,rocco2017convolutional,han2017scnet} while scores marked with an asterisk (*) are drawn from the reproduced models by released official codes.
The PCK scores marked with a dagger ($^\dagger$) are borrowed from~\cite{han2017scnet} measuring them with respect to the entire image size instead of the bounding box size.
Note that the PCK score with respect to the bounding box size is more conservative than the one with respect to the image size, resulting in lower scores. For example, PCK scores of A2Net on PF-PASCAL measured by image sizes are 0.59 (affine), 0.65 (affine+TPS) and 0.71 (affine+TPS; ResNet101). \\ 
}

\label{tab:main_results}
\begin{tabular}{
@{}C{0.6cm}@{}|@{}C{0.3cm}@{}@{}p{5.5cm}@{}|@{}C{2.3cm}@{}|@{}C{2.3cm}@{}
}
&&\multirow{2}{*}{Models}                             & \multicolumn{2}{c}{PCK ($\alpha=0.1$)}      \\
&&                                                    & PF-WILLOW    & PF-PASCAL     \\ \hline\hline
\multirow{7}{*}{\rb{Hand-crafted~}}&&DeepFlow~\cite{revaud2016deepmatching}              & 0.20         & 0.21          \\
&&GMK~\cite{duchenne2011graph}                        & 0.27         & 0.27          \\
&&SIFTFlow~\cite{liu2016sift}                         & 0.38         & 0.33          \\
&&DSP~\cite{kim2013deformable}                        & 0.29         & 0.30          \\
&&ProposalFlow (NAM)~\cite{ham2016proposal}           & 0.53         &  --           \\
&&ProposalFlow (PHM)~\cite{ham2016proposal}           & 0.55         &  --           \\
&&ProposalFlow (LOM)~\cite{ham2016proposal}           & 0.56         & 0.45          \\ \hline
\multirow{4}{*}{\rb{Self Sup.~~~}}&&GeoCNN (affine)~\cite{rocco2017convolutional}       & 0.49         & 0.39*         \\
&&GeoCNN (affine+TPS)~\cite{rocco2017convolutional}   & 0.56         & 0.50*         \\
&&A2Net (affine)                                       & 0.52         & 0.46          \\
&&A2Net (affine+TPS)                                   & 0.63    & 0.54     \\ 
&&\bf{A2Net (affine+TPS; ResNet101)}                   & \bf{0.68}    & \bf{0.59}     \\ \hline
\multirow{5}{*}{\rb{{Supervised}~}}&&UCN~\cite{choy2016universal}                        & 0.42$^\dagger$         & 0.56$^\dagger$          \\
&&FCSS~\cite{kim2017fcss}                             & 0.58        & --          \\
&&SCNet-A~\cite{han2017scnet}                         & 0.73$^\dagger$         & 0.66$^\dagger$          \\ 
&&SCNet-AG~\cite{han2017scnet}                        & 0.72$^\dagger$         & 0.70$^\dagger$          \\ 
&&SCNet-AG+~\cite{han2017scnet}                       & 0.70$^\dagger$         & 0.72$^\dagger$          \\ 
\hline
\end{tabular}
\end{table}

\subsection{Results}
\subsubsection{Comparisons to other models} 
Table~\ref{tab:main_results} shows the comparative results on both PF-WILLOW and PF-PASCAL benchmarks.
It includes (i) previous methods using hand-crafted features: DeepFlow~\cite{revaud2016deepmatching}, GMK~\cite{duchenne2011graph}, SIFTFlow~\cite{liu2016sift}, DSP~\cite{kim2013deformable}, and ProposalFlow~\cite{ham2016proposal}, (ii) self-supervised alignment methods: GeoCNN~\cite{rocco2017convolutional} and the proposed attentive alignment network (A2Net),  (iii) supervised methods: UCN~\cite{choy2016universal}, FCSS~\cite{kim2017fcss}, and SCNet~\cite{han2017scnet}. 
Note that the supervised methods are trained with either a weakly or strongly annotated data and that many of their PCKs are measured under a different criterion that are not directly comparable to the other scores. 
By contrast, our method is only trained using synthetic data with self-supervision.
As shown in Table~\ref{tab:main_results}, the proposed method substantially outperforms all the other methods that are directly comparable. 
Using VGG-16 feature extractor, the proposed method improves 12.5 \% and {8} \% of PCK over the non-attentive alignment method~\cite{rocco2017convolutional} on PF-WILLOW and PF-PASCAL, respectively.
This reveals the effect of the proposed attention model for semantic alignment. 
The quality of the model is further improved when incorporated with a more advanced feature extractor such as ResNet101.
It is notable that the proposed model outperforms some of supervised methods, UCN~\cite{choy2016universal} and FCSS~\cite{kim2017fcss}, while it is trained without any real datasets.


\begin{table}[t]
\centering
\caption{
PCKs of ablations on PF-WILLOW trained with PASCAL VOC 2011. 
Scores of GeoCNN are obtained from the code released by the authors.
The numbers of network parameters exclude the feature extractors since all models share the same feature extractor.
}
\label{tab:ablation}
\begin{tabular}{
@{}C{0.2cm}@{}
@{}p{4cm}@{}|@{}C{2.4cm}@{}|@{}C{1.5cm}@{}@{}C{1.5cm}@{}@{}C{2.1cm}@{}
}
&Models                     &\# of params       & Affine    & TPS       & Affine+TPS    \\ \hline\hline
&GeoCNN~\cite{rocco2017convolutional} & 1.63M (x1.7)      & 0.430     & 0.539     & 0.560         \\
&GeoCNN+Attention           & 1.41M (x1.5)      & 0.423     & 0.478     & 0.476         \\
&GeoCNN+OACK                & 1.12M (x1.2)      & 0.491     & 0.555     & 0.609         \\
&Attention+OACK (A2Net)      & \bf{0.95M (x1.0)} & \bf{0.521}& \bf{0.563}& \bf{0.626}    \\
\hline
\end{tabular}
\end{table}

\begin{table}[t]
\centering
\caption{PCKs of affine models on PF-WILLOW with different training datasets: PASCAL VOC 2011 and Tokyo Time Machine. Scores for GeoCNN are brought from \cite{rocco2017convolutional}.}
\label{tab:generalization}
\begin{tabular}{
@{}C{0.2cm}@{}
@{}p{2.2cm}@{}|@{}C{3.2cm}@{}|@{}C{3.2cm}@{}
}
&Models                     & PASCAL VOC 2011   & Tokyo Time Machine    \\ \hline\hline
&GeoCNN~\cite{rocco2017convolutional} & 0.45              & 0.49                  \\
&A2Net                       & \bf{0.52}         & \bf{0.51}             \\
\hline
\end{tabular}
\end{table}

\subsubsection{Ablation study}

As our proposed model combines two distinct techniques we perform ablation studies to demonstrate their effects.
We mainly compare the proposed model to GeoCNN as it directly predicts the global transformation parameters using the correlation layer.
To see the effect of the proposed OAC kernels, we build a model, referred to as GeoCNN+OACK, by replacing the correlation layer of GeoCNN with the OAC kernels.
As shown in Tabel~\ref{tab:ablation}, the use of the OAC kernels already improves the performances of GeoCNN for all three versions.
Moreover, the OAC kernels reduce the number of parameters in the network since it uses dense representations of local transformations allowing channel compression.
Applying attention process on top of correlation layer (GeoCNN+Attention) drops the performance.
This is because the correlation map does not encode local transformations in a translation invariant representations.
On the other hand, the attention process with the OAC kernels, which is the proposed model, further improves the performances as the distracting regions can be suppressed during the transformation estimation thanks to the local transformation feature map obtained by the OAC kernels.
It is also notable that applying the attention process reduces the number of model parameters because the model does not need extra layers that combine all local information to produce the global estimation; instead, the models simply aggregate local features with attention distribution.
This additional parameter reduction results in 70 \% fewer parameters than GeoCNN while the models maintain superior performance improvements.

\subsubsection{Sensitivity to training datasets}
While both our model and GeoCNN are generally applicable to any image datasets, we experiment the sensitivity of the models to changing training datasets.
We train both models with the affine transformation on another image dataset, called Tokyo Time Machine~\cite{arandjelovic2016netvlad}, using the same synthetic generation process, and show how much the performances change depending on the datasets.
Table~\ref{tab:generalization} shows that the proposed model is less dependent on the choice of the training dataset compared to GeoCNN.

\begin{figure}[t]
    \centering
    \includegraphics[width=1\linewidth]{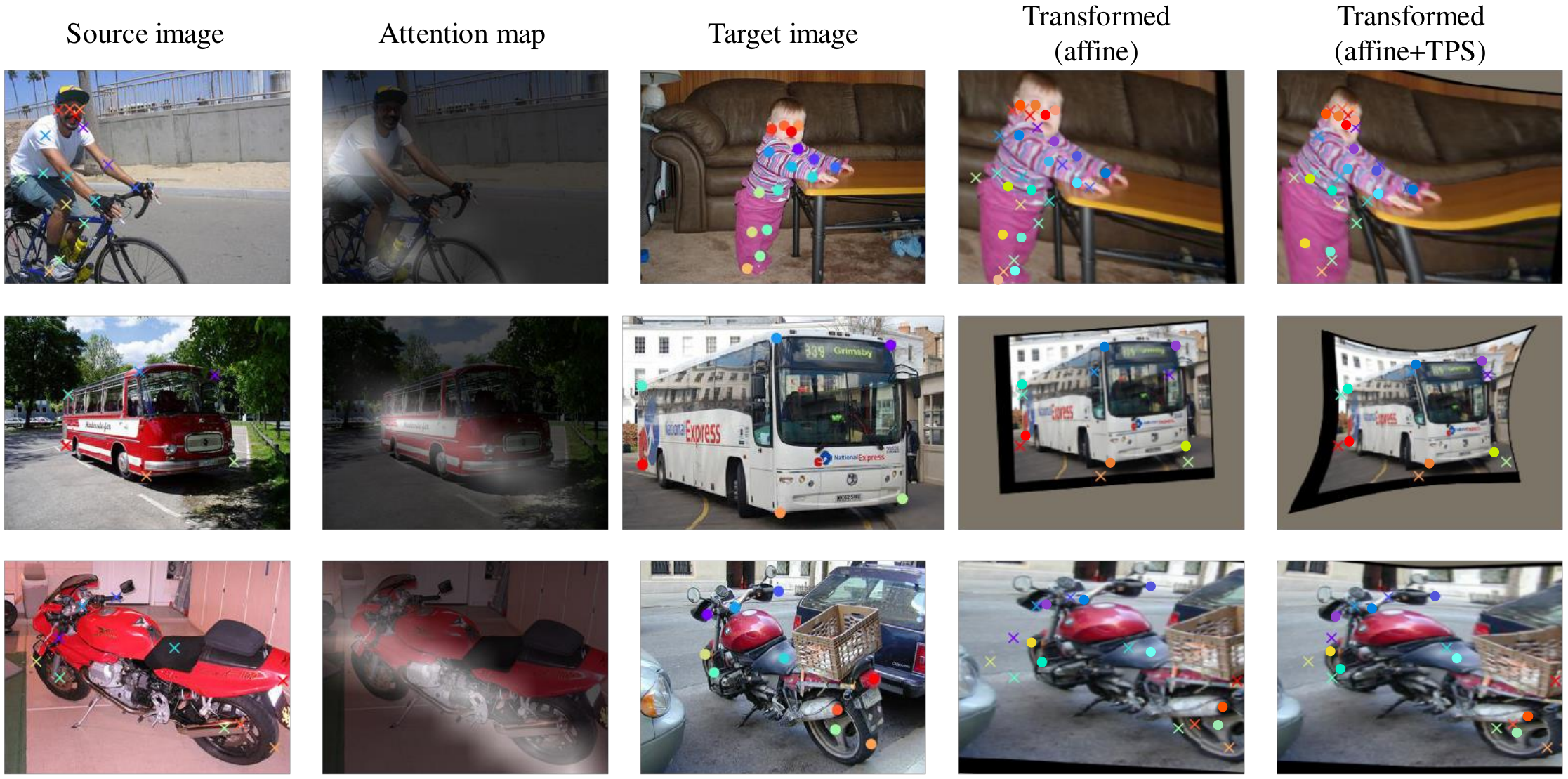}
    \caption{
    Qualitative results of the attentive semantic alignment.
    Each row shows an example of PF-PASCAL benchmark.
    Given the source and target images shown in first and third columns, we visualize the attention maps of the affine model (second column), the transformed image by the affine model (fourth column) and the final transformed image by the affine+TPS model (last column).
    Since the models learn inverse transformation, the target image is transformed toward the source image while the attention distribution is drawn over the source image.
    The model attends to the objects to match and estimates dense correspondences despite intra-class variations and background clutters.
    }
    \label{fig:qualitative}
\end{figure}

\subsubsection{Qualitative results with attention visualizations}
Fig.~\ref{fig:qualitative} presents some qualitative examples of our model on PF-PASCAL.
In our experimental setting, the models learn to predict inverse transformation.
Therefore, we transform the target image toward the source image using the estimated inverse transformation whereas the attention distribution is drawn over the source image.
The proposed model attends to the target objects with other regions suppressed and predicts the global transformation based on reliable local features.
The model estimates the transformation despite large intra-class variations such as an adult vs. a kid.

\begin{figure}[t]
    \centering
    \includegraphics[width=0.9\linewidth]{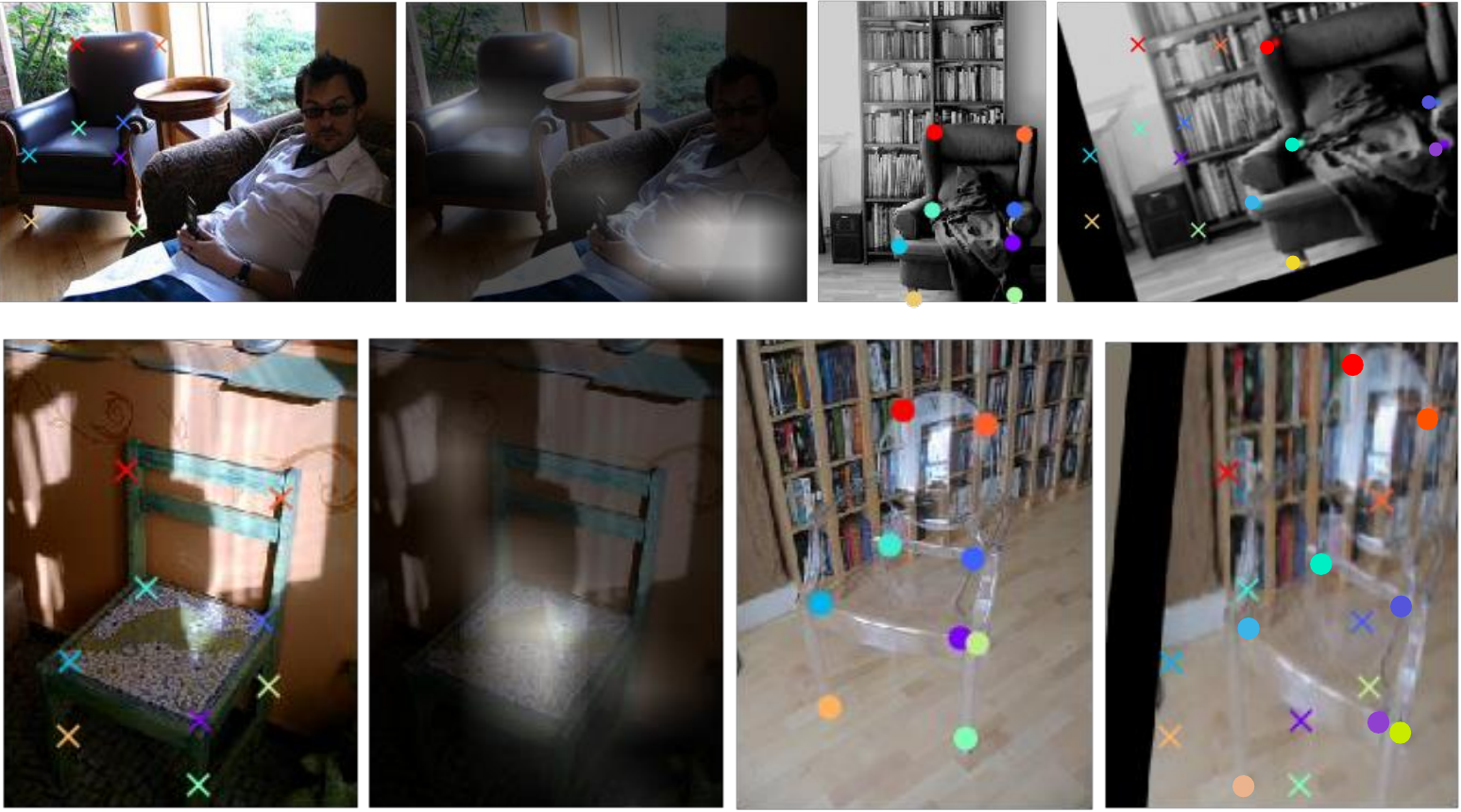}
    \caption{
    Some failure cases of the proposed model with the affine transformation.
    Each row shows an example of PF-PASCAL.
    Each example contains 1) source image, 2) source image masked by attention distribution, 3) target image and 4) target image transformed by the predicted affine parameters.
    Even though the model attends to the matching objects, the model fails to find the correct correspondences due to multiple objects of the same class causing ambiguity or hard examples that are difficult to visually percept.
    }
    \label{fig:failure}
\end{figure}

We also investigate some failure cases of the proposed model in Fig.~\ref{fig:failure}.
The model is confused when there are multiple objects of the same class in an image or have a large obstacles occluding the matching objects.
Also, objects in some examples are hard to visually recognize and lead mismatches.
For instance, the model fails to correctly match a wooden chair to a transparent chair although the model attends to the correct region in the second example of Fig.~\ref{fig:failure}.
It is challenging even for human to recognize the transparent chair and its corresponding key points.

\section{Conclusion}
\label{sec:conclusion}

We propose a novel approach for semantic alignment.
Our model facilitates an attention process to estimate global transformation from reliable local transformation features by suppressing distracting features.
We also propose offset-aware correlation kernels that reorder correlations of feature pairs and produce a dense feature representation of local transformations.
The experimental results show the attentive model with the proposed kernels achieves the state-of-the-art performances with large margins over previous methods on the PF-WILLOW and PF-PASCAL benchmarks.

\section*{Acknowledgement}
This research was supported by Next-Generation Information Computing Development Program (NRF-2017M3C4A7069369) and Basic Science Research Program (NRF-2017R1E1A1A01077999) through the National Research Foundation of Korea (NRF) funded by the Ministry of Science, ICT, and Institute for Information \& communications Technology Promotion (IITP) funded by the Korea government (MIST) (No. 2017-0-01778, Develoment of Explainable Human-level Deep Machine Learning Inference Framework).

\bibliographystyle{splncs}
\bibliography{egbib}

\clearpage

\beginsupplement
\section*{\em Supplementary Material}
\begin{appendix}
\section{Evaluation on Other Datasets}
We have evaluated the proposed model on two more datasets: Taniai's dataset~\cite{taniai2016joint} and Caltech~101~\cite{fei2006one}.

\subsubsection{Results on  on Taniai's dataset}
Taniai's dataset~\cite{taniai2016joint} contains 400 image pairs in three subsets: FG3DCar (195 pairs from \cite{lin2014jointly}), JODS (81 pairs from \cite{rubinstein2013unsupervised}) and PASCAL (124 pairs from \cite{hariharan2011semantic}).
Each image pair in this dataset is annotated with dense flows based on key points.
Following \cite{taniai2016joint}, we measure flow accuracy which is the percentage of correctly transferred flows.
Each flow is considered correct if the distance between the estimated flow and the ground-truth flow is less than 5 pixels.

Table~\ref{tab:tss} shows the results on Taniai's dataset.
Our model shows higher flow accuracies over all other models on every subset of the dataset.
Especially, we emphasize that the proposed model shows significant gains over GeoCNN which uses a single end-to-end neural network for global transformation estimation as in our model.

\subsubsection{Results on Caltech-101}
Caltech-101~\cite{fei2006one} consists of 1515 image pairs of 101 object classes.
Unlike other datasets, these image pairs are not annotated with dense correspondences.
Instead, we utilize the annotated segmentation masks for the evaluation.
Following~\cite{rocco2017convolutional}, we measure label transfer accuracy (LT-ACC) and intersection over union (IoU) of transformed masks.
LT-ACC measures the accuracy of pixel-level segmentation labels between the transformed masks and the ground-truth masks, and IoU measures the intersection over union of those masks. 
We believe that localization error (LOC-ERR) used in \cite{rocco2017convolutional,ham2016proposal,han2017scnet} is not appropriate for evaluating dense correspondences as it measures the error based on the bounding box information extracted from masks losing the precise details, but we report this score for completeness.

Table~\ref{tab:caltech} shows the results on Caltech-101.
Our model shows the best score in IoU and LT-ACC over all the methods.

\begin{table}[t]
\centering
\caption{
Flow accuracies with error threshold with 5 pixels on Taniai's dataset.
Scores of DCTM is obtained with VGG-16 feature extractor as in our method since the FCSS feature extractor in the original paper is trained with a weak supervision.
}
\label{tab:tss}
\begin{tabular}{
@{}C{0.2cm}@{}
@{}p{3.5cm}@{}|@{}C{1.6cm}@{}@{}C{1.6cm}@{}@{}C{1.6cm}@{}|@{}C{1.6cm}@{}
}
&Models                                 & FG3DCar   & JODS      & PASCAL    & Average   \\ \hline\hline
&DFF~\cite{yang2014daisy}               & 0.50      & 0.30      & 0.22      & 0.31      \\
&DSP~\cite{hur2015generalized}          & 0.49      & 0.47      & 0.38      & 0.45      \\
&SIFT Flow~\cite{liu2016sift}           & 0.63      & 0.51      & 0.36      & 0.50      \\
&Zhou \etal~\cite{zhou2016learning}     & 0.72      & 0.51      & 0.44      & 0.56      \\
&Taniai \etal~\cite{taniai2016joint}    & 0.83      & 0.60      & 0.48      & 0.64      \\
&Proposal Flow~\cite{ham2016proposal}   & 0.79      & 0.65      & 0.53      & 0.66      \\
&DCTM (VGG-16)~\cite{kim2017dctm}       & 0.79      & 0.61      & 0.53      & 0.63      \\ \hline
&GeoCNN~\cite{rocco2017convolutional}   & 0.85      & 0.64      & 0.53      & 0.67      \\
&Proposed                               & \bf{0.87} & \bf{0.67} & \bf{0.55} & \bf{0.70} \\
\hline
\end{tabular}
\end{table}

\begin{table}[t]
\centering
\caption{
Results on Caltech-101. 
*The scores for GeoCNN are reproduced using publicly released code by the authors and show slightly lower LT-ACC.
}
\label{tab:caltech}
\begin{tabular}{
@{}C{0.2cm}@{}
@{}p{2.8cm}@{}|@{}C{1.6cm}@{}@{}C{1.6cm}@{}@{}C{1.6cm}@{}
}
&Models                                     & LT-ACC    & IoU       & LOC-ERR   \\ \hline\hline
&DeepFlow~\cite{revaud2016deepmatching}     & 0.74      & 0.40      & 0.34      \\
&GMK~\cite{duchenne2011graph}               & 0.77      & 0.42      & 0.34      \\
&DSP~\cite{hur2015generalized}              & 0.77      & 0.47      & 0.35      \\
&SIFT Flow ~\cite{liu2016sift}              & 0.75      & 0.48      & 0.32      \\
&Proposal Flow~\cite{ham2016proposal}       & 0.78      & 0.50      & \bf{0.25}      \\ \hline
&GeoCNN~\cite{rocco2017convolutional}*      & 0.79      & 0.56      & 0.28      \\
&Proposed                                   & \bf{0.80}      & \bf{0.57} & 0.28      \\
\hline
\end{tabular}
\end{table}

\section{Visualizations of OAC Kernel Weights}
We visualize the learned weights of OAC kernels in the proposed model.
Fig.~\ref{fig:kernels} presents three examples of kernel weights arranged by the offsets along with X and Y axes. 
Kernel weights are focused on a certain region with similar offsets capturing a displacement to that direction.

\begin{figure}[h]
\centering

\includegraphics[width=0.22\linewidth] {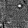}
~~~~~~
\includegraphics[width=0.22\linewidth] {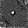}
~~~~~~
\includegraphics[width=0.22\linewidth] {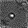}

\caption{
Visualizations of some offset-aware correlation kernels.
Each kernel captures displacements by learning pointy weights focused on a region with similar offsets.
}
\label{fig:kernels}
\end{figure}

\section{More Qualitative Results}

\begin{figure}[h]
\centering

\includegraphics[width=1\linewidth] {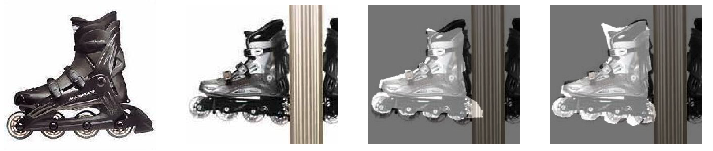}
\includegraphics[width=1\linewidth] {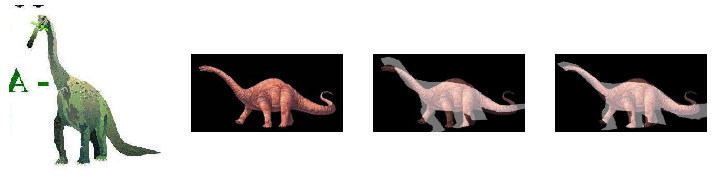}
\includegraphics[width=1\linewidth] {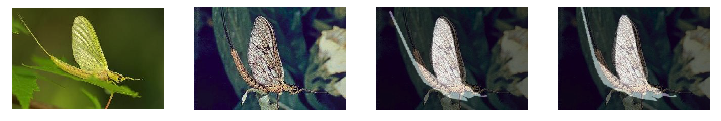}
\includegraphics[width=1\linewidth] {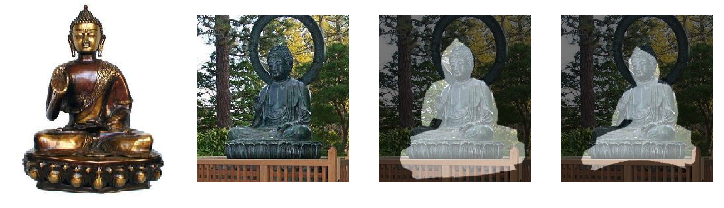}

\caption{
Qualitative examples from Caltech-101.
Each row presents different examples including source image (first column), target image (second column), target image with affine transformed source mask (third column), and target image with TPS transformed source mask (fourth column).
}
\label{fig:qualitative_caltech}
\end{figure}

We present additional qualitative results from Caltech-101 dataset.
In Fig.~\ref{fig:qualitative_caltech}, segmentation masks of source images are transformed and visualized on target images.

\begin{table}[t]
    \centering
    \vspace*{-0.3cm}
    \caption{
        IoU on different test subsets of Caltech-101. See text for detail. 
    }
    \label{tab:result_imgnet}
    \begin{tabular}{@{}C{3.3cm}@{}|@{}C{1.3cm}@{}@{}C{1.3cm}@{}@{}C{1.9cm}@{}}
                            & Affine    & TPS   & Affine+TPS\\ \hline
        ImageNet objects    & 0.54      & 0.56  & 0.58      \\
        Other objects       & 0.52      & 0.53  & 0.56      \\ \hline
        All objects         & 0.53      & 0.55  & 0.57      \\
    \end{tabular}
    \vspace*{-0.3cm}
\end{table}

\section{Effect of Pretrained Network}
To investigate the importance of the pretrained recognition capability of the feature extractor, we evaluated our model on different test subsets using Caltech101 dataset.
The first subset consists of test pairs containing objects that appear in ImageNet, which is used to pretrain our feature extractor, and the second subset contains pairs whose objects are unseen at the pretraining.
We use Caltech101 dataset for this experiment because almost all objects in PF-WILLOW and PF-PASCAL also appear in ImageNet. 
Table~\ref{tab:result_imgnet} summarizes IoU results of our model on the test subsets.
We can see that pairs of objects that appear in ImageNet show slightly better performance than the other pairs.
However, the gap is relatively small, indicating that the model is still capable of predicting transformations of objects that are unseen during the pretraining.
We conjecture that this is because our alignment prediction (1) relies on the feature correlations, which are less class-specific than the raw features themselves, and (2) uses intermediate-level visual features (extracted at \texttt{pool4}), which are rather class-agnostic compared to higher-level features.


\section{Experiments on Cross-class Matching}
We also conduct the cross-class matching experiments as done in the work of \cite{novotny2017anchornet}.
As the authors have neither provided their code online nor responded to our request yet, we did our best to reproduce the experimental settings of \cite{novotny2017anchornet} for ourselves.
However, some of the descriptions are unclear and thus there might be some differences.
In the experiment of measuring weighted IoU (wIoU) of part segmentations for cross-class objects on the PASCAL Parts dataset, where the best model in \cite{novotny2017anchornet} produces wIoU of 37.5 \%, our method shows a competitive wIoU of 35.0 \%. 
As mentioned earlier, this may be due to the use of the feature correlations, rather than the raw features themselves, in alignment prediction. If the features of cross-class objects share similar representations, our method is still capable of predicting the alignment between them.
\end{appendix}
\end{document}